\pdfoutput=1

\PassOptionsToPackage{table}{xcolor}

\documentclass[11pt]{article}

\usepackage[]{acl}

\usepackage{times}
\usepackage{latexsym}

\usepackage[T1]{fontenc}

\usepackage[utf8]{inputenc}

\usepackage{microtype}

\usepackage{amsmath,graphicx,booktabs,amssymb}
\usepackage{hyperref}

\usepackage{wasysym,xspace}
\usepackage{comment}
\usepackage{multirow}

\PassOptionsToPackage{table}{xcolor}

\usepackage[table]{xcolor}

\title{Explaining Speech Classification Model Predictions} 
\title{Explaining Speech Classification Models via Input Perturbation}
\title{What Did You Just Say? Explaining Speech Classification Models via Input Perturbation}
\title{Explaining Speech Classification Models via Word-Level \\ Audio Segments and Paralinguistic Features}

\newcommand{\bocconi}{$^{\heartsuit}$}
\newcommand{\polito}{$^{\clubsuit}$}
\newcommand{\repo}{\url{https://github.com/elianap/SpeechXAI}\xspace}

\author{Eliana Pastor\polito, Alkis Koudounas\polito, Giuseppe Attanasio\bocconi, Dirk Hovy\bocconi, Elena Baralis\polito\\ \\
 \polito~Politecnico di Torino, Turin, Italy\\
 \bocconi~Bocconi University, Milan, Italy \\
 \\
 {
 \tt \{eliana.pastor,alkis.koudounas,elena.baralis\}@polito.it} \\
    \tt \{giuseppe.attanasio3,dirk.hovy\}@unibocconi.it}

\begin{document}
\maketitle
\begin{abstract}

Recent advances in eXplainable AI (XAI) have provided new insights into how models for vision, language, and tabular data operate. However, few approaches exist for understanding speech models. Existing work focuses on a few spoken language understanding (SLU) tasks, and explanations are difficult to interpret for most users.
We introduce a new approach to explain speech classification models. We generate easy-to-interpret explanations via input perturbation on two information levels. 1) Word-level explanations reveal how each word-related audio segment impacts the outcome. 2) Paralinguistic features (e.g., prosody and background noise) answer the counterfactual: ``What would the model prediction be if we edited the audio signal in this way?''
We validate our approach by explaining two state-of-the-art SLU models on two speech classification tasks in English and Italian. Our findings demonstrate that the explanations are faithful to the model's inner workings and plausible to humans.
Our method and findings pave the way for future research on interpreting speech models.
\smallskip
\\
\textit{Note: This preprint documents our approach and preliminary results. We are working on expanding the evaluations and discussions.}
\end{abstract}

\section{Introduction}
\label{sec:intro}

Recently, several eXplainable AI (XAI) techniques have been proposed to gain insights into how models get to their outputs. 
Seminal work in computer vision used gradients \citep[\textit{inter alia}]{simonyan2013deep,sundararajan2017axiomatic,selvaraju2022grad} or input perturbation \citep{zeiler2013visualizing} to build input saliency maps, i.e., visual artifacts to highlight the most relevant parts for the prediction. Similar solutions have also been proposed to explain language \citep[\textit{inter alia}]{ribeiro2016should,sanyal-ren-2021-discretized,jacovi-etal-2021-contrastive} and tabular \citep{lundberg_unified_2017} models.

\begin{figure}
    \centering
    \includegraphics[width=.95\linewidth]{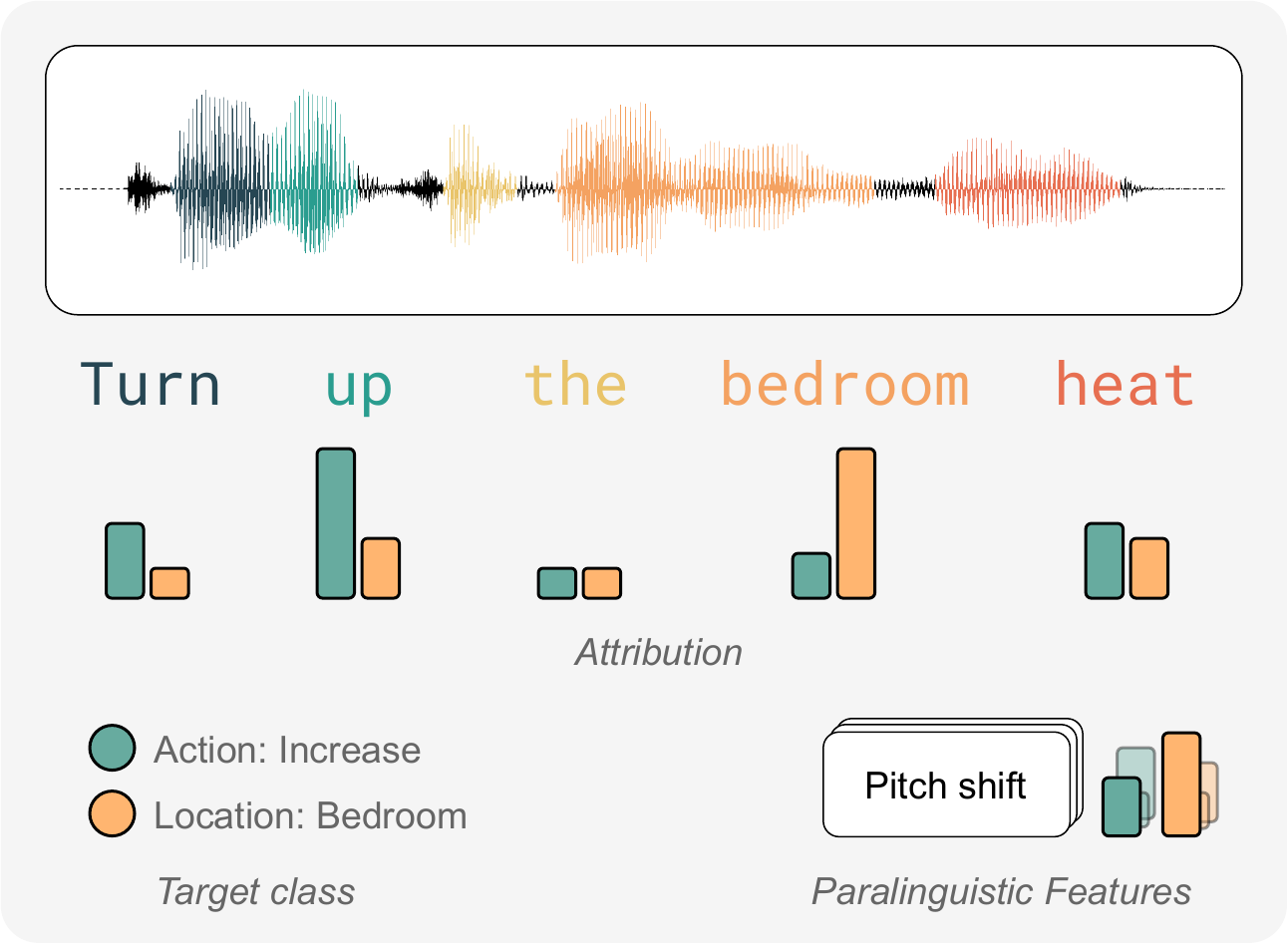}
    \caption{Explanation with word-level and paralinguistic attributes for a sample in Fluent Speech Commands \citep{LugoschRITB19}. Word-level audio-transcript alignment represented through color. Word-level attributions to explain the \textit{Increase} (green, left boxes) and \textit{Bedroom} (orange, right) target classes.}
    \label{fig:explanation}
\end{figure}

And while there is significant progress in explaining model predictions for image, text, and structured data models, explanations for Spoken Language Understanding (SLU) models remain largely unexplored.
Speech data consists of both explicit content and discrete words, but also acoustic features, linguistic variations, and paralinguistic cues, making it more complex to decipher each element's contribution to the model predictions.
Existing approaches use frequency features, e.g., spectrogram segments \citep{becker2018interpreting, frommholz2023xai}. However, spectrograms are  difficult to interpret for most humans. \citet{wu2023can} have instead proposed identifying time segments, e.g., those corresponding to relevant phonemes. However, meaningful, phoneme-level explanations are fine-grained and only serve a limited number of tasks like Automatic Speech Recognition (ASR) or Phoneme Recognition. They fail to capture more interpretable word-level attribution needed for semantically-intensive tasks such as Speech Classification.
Moreover, these methods \textit{entirely overlook} any paralinguistic aspects, e.g., prosody or channel noise, which carry information. 



We propose a new approach to explaining speech models, producing easy-to-interpret explanations including paralinguistic features. 
We base our approach on input perturbation, an established XAI method. Our explanations provide insights on two different but complementary levels: The uttered content and paralinguistic features.

To quantify the contribution of each part of the utterance, we compute word-level attribution scores as follows. First, we align the audio signal to its transcript and get word-level timestamps. Then, we use these timestamps to iteratively mask audio segments. Finally, we estimate word-level contributions as the difference in the model's output between the original signal and the masked one.
We follow a similar perturbation-based approach to measure the contribution of paralinguistic aspects. Given an input utterance, we transform the raw audio signal and measure the effect on the model's prediction. We perturb pitch to account for prosody, and audio stretching, background noise, and reverb levels for channel-related aspects. 
Figure~\ref{fig:explanation} shows a sample explanation.

We test our approach by explaining wav2vec-2.0 \citep{wav2vec2} and XLS-R \citep{babu2021xls}, two state-of-the-art SLU models, on two datasets for Intent Classification and one for Emotion Recognition in English and Italian.  
We assess the quality of our explanations under the faithfulness and plausibility paradigms \citep{jacovi-goldberg-2020-towards}.
Our experimental results demonstrate that the explanations are faithful to the model's inner workings and plausible to humans. 

\paragraph{Contributions.}
We introduce a new method for explaining speech classification models. Using word-level audio segments and paralinguistic features, it generates easy-to-interpret visualizations that are faithful and plausible across two models, languages, and tasks. 
We release the code at \repo to encourage future research at the intersection of SLU and interpretability. 

\section{Methodology}
\label{sec:method}

We generate explanations by assigning a single numerical attribution score to each uttered word (\S\ref{ssec:word_level_attr}) and paralinguistic feature (\S\ref{ssec:paralinguistic_attr}). Each score is generated via input perturbation and quantifies the contribution the entity (either a word or a paralinguistic feature) had in predicting a given target class. 

\subsection{Word-level Audio Segment Attribution}
\label{ssec:word_level_attr}

We compute word-level contribution in two steps. First, we perform a word-level audio-transcript alignment. In practice, we extract beginning and ending timestamps for each uttered word. If no transcript or timestamp is available, we use WhisperX \citep{bain2022whisperx} to generate it along with the word-level timestamps. The resulting timestamps define a set of audio segments corresponding to words in the time domain.
See Figure~\ref{fig:explanation} (top) for an example.

Second, we compute each segment's contribution by masking it and measuring how the model's prediction changes. 
More formally, let $x$ be an utterance and let $\{x_1, .., x_n\}$ the set of $n$ word-level audio segments within.
Consider a speech classification model $f$ applied for tasks such as intent classification or emotion recognition. 
Let $f(y=k|x)$ be the output probability of the model $f$ for class $k$  given the input utterance $x$. 
We define the relevance $r(x_i) \in \mathbb{R}$ of each segment $x_i$ to the model's prediction for a target class $k$ as:
\begin{equation}
r(x_i) = f(y=k|x)-f(y=k|x \setminus x_i)
\end{equation}
where $x \setminus x_i$ refers to the utterance when the segment $x_i$ is masked. Following \citet{wu2023can}, we mask out segments by zeroing the corresponding samples in the time domain.


Higher values for $r(x_i)$ indicate greater relevance of the segments to the prediction. A positive score indicates that the segment contributes positively to the probability of belonging to a specific class, while a negative score suggests that the segment may ``push'' the prediction toward another class. 
See Figure~\ref{fig:explanation} (middle) for an example.

\subsection{Paralinguistic Attributions}
\label{ssec:paralinguistic_attr}

Speech includes not only the semantic information conveyed by words but also additional paralinguistic information communicated through the speaker's voice or from external conditions, such as pitch, speaking rate, and background noise levels. We investigate the relevance of paralinguistic features by introducing ad hoc perturbations of the utterances and studying the resulting changes in class prediction probabilities.

Let $p(x)$ be a paralinguistic feature of interest of utterance $x$. For example, it can correspond to the pitch of the utterance.
We transform $x$ into $\widetilde{x}$ such that the value of feature $p(\widetilde{x})$ varies from $p(x)$. Rather than a random perturbation, we control the induced transformation so that it is interpretable, and we can trace back the impact to feature $p$.
For instance, we may increase the pitch. 

We consider a series of transformation $\widetilde{X}_p$ = \{$\widetilde{x}_1, .., \widetilde{x}_t $\}  to study the impact of changing the paralinguistic feature $p$ on the model's predictions.
We compute the relevance of $p(x)$ as follows.
\begin{equation}
r(p(x)) = f(y=k|x) - \frac{1}{|X|} \sum_{\widetilde{x}  \in \widetilde{X} } f(y=k|\widetilde{x})     
\end{equation}
The term $\frac{1}{|X|} \sum_{\widetilde{x}  \in \widetilde{X} }  f(y=k|\widetilde{x})  $ represents the average change in the prediction probability when perturbing $p(x)$.
In addition, we visualize the terms $f(y=k|x) - f(y=k|\widetilde{x})$ in a heatmap representation to visualize the impact of each perturbation. Heatmaps provide an intuitive way to observe the changes in prediction probabilities as we vary the paralinguistic features.

\section{Experiments}
\label{sec:results}

\subsection{Experimental Setting} 


\paragraph{Paralinguistic Features.} In the experiments, we consider transformations of the pitch, time stretching, the introduction of background white noise, and of reverberation.
We describe the libraries adopted for the transformations in our repository.

\paragraph{Datasets.} 
We evaluate our explanation on three publicly available datasets and two tasks: \textsc{Fluent Speech Commands}~\citep[FSC;][]{LugoschRITB19} and the Italian Intent Classification Dataset~\citep[ITALIC;][]{koudounas2023italic} datasets for Intent Classification (IC) task and the \textsc{IEMOCAP}~\citep{IEMOCAP} for Emotion Recognition (ER).
\textsc{FSC} is a widely utilized benchmark dataset for the IC task. 
Its test set comprises 3793 audio samples, each characterized by three slots — action, object, and location — whose combination defines the intent. 
\textsc{ITALIC} is an intent classification dataset for the Italian language.
The dataset includes 60 intents, and the test set consists of 1441 samples.
We use the ``Speaker'' setup, wherein the utterances of each speaker belong to a single set among the train, validation, and test.
\textsc{IEMOCAP} is a dataset for the ER task annotated with
emotion labels (i.e., happiness, anger, sadness, frustration, and neutral state).
It consists of recorded interactions between pairs of actors engaged in scripted scenarios involving ten actors. Among its five sessions, we consider 
Session `1', consisting of 942 utterances. 

\paragraph{Models.} We consider the monolingual wav2vec 2.0 base~\citep{wav2vec2} for \textsc{FSC} and \textsc{IEMOCAP}. We use the public fine-tuned checkpoints~\cite{yang2021superb}. We use the multilingual XLS-R~\citep{babu2021xls} for ITALIC and its fine-tuned checkpoints~\citep{koudounas2023italic}.

\subsection{Qualitative evaluation}

In this section, we show how our explanation method reveals the reasons behind a model prediction from the perspective of an \textit{individual} prediction and \textit{globally} across the entire dataset. 

\paragraph{Individual level.} Consider the \textsc{FSC} dataset and wav2vec 2.0 base fine-tuned-model. 
For a specific utterance with transcription `Turn up the bedroom heat', the model correctly predicts \textit{increase} as the action, \textit{heat} as the object, and \textit{bedroom} as the location, fully identifying the intent.
We may wonder: Is it correct for the right reasons? Which are the paralinguistic features whose change would impact the predictions? Our approach answers these questions.

\begin{table}[!t]
\centering
\setlength{\tabcolsep}{3pt}
\scalebox{0.91}{
\begin{tabular}{c|ccccc}
\hline
{\color[HTML]{000000} \textbf{}}    & {\color[HTML]{000000} \textbf{Turn}}                  & {\color[HTML]{000000} \textbf{up}}                    & {\color[HTML]{000000} \textbf{the}}                   & {\color[HTML]{000000} \textbf{bedroom}}              & {\color[HTML]{000000} \textbf{heat.}}                \\ \hline
{\color[HTML]{000000} act=increase} & \cellcolor[HTML]{ECC3C6}{\color[HTML]{000000} 0.250}  & \cellcolor[HTML]{E58D94}{\color[HTML]{000000} 0.545}  & \cellcolor[HTML]{ECC2C5}{\color[HTML]{000000} 0.260}  & \cellcolor[HTML]{EFD9DA}{\color[HTML]{000000} 0.139} & \cellcolor[HTML]{F2EEEE}{\color[HTML]{000000} 0.021} \\
{\color[HTML]{000000} obj=heat}     & \cellcolor[HTML]{F1F1F2}{\color[HTML]{000000} 0} & \cellcolor[HTML]{F1F1F2}{\color[HTML]{000000} 0} & \cellcolor[HTML]{F1F1F2}{\color[HTML]{000000} 0} & \cellcolor[HTML]{F2EFEF}{\color[HTML]{000000} 0.014} & \cellcolor[HTML]{E58C93}{\color[HTML]{000000} 0.550} \\
{\color[HTML]{000000} loc=bedroom}  & \cellcolor[HTML]{F2F1F1}{\color[HTML]{000000} 0.002}  & \cellcolor[HTML]{F2F1F1}{\color[HTML]{000000} 0.006}  & \cellcolor[HTML]{F0E1E2}{\color[HTML]{000000} 0.087}  & \cellcolor[HTML]{DA3B46}{\color[HTML]{000000} 0.997} & \cellcolor[HTML]{EAB6BA}{\color[HTML]{000000} 0.323} \\ \hline
\end{tabular}
}
\caption{Example of word-level audio segment explanation; \textsc{FSC} dataset. The higher the value, the more the audio segment is relevant for the prediction.}
\label{tab-example-fsc-word}
\end{table}
\begin{table}[!t]
\centering
\setlength{\tabcolsep}{2pt}
\scalebox{0.9}{
\begin{tabular}{c|cccccc}
\hline
 & \multicolumn{2}{c}{\textbf{pitch}} & \multicolumn{2}{c}{\textbf{stretch}} &  &  \\
\textbf{} & \textbf{down} & \textbf{up} & \textbf{down} & \textbf{up} & \multirow{-2}{*}{\textbf{reverb}} & \multirow{-2}{*}{\textbf{noise}} \\ \cline{1-5}
act=increase & \cellcolor[HTML]{F2F1F1}{\color[HTML]{000000} 0} & \cellcolor[HTML]{F2EFEF}{\color[HTML]{000000} 0.01} & \cellcolor[HTML]{EED0D2}{\color[HTML]{000000} 0.19} & \cellcolor[HTML]{F2EBEB}{\color[HTML]{000000} 0.04} & \cellcolor[HTML]{E06A73}{\color[HTML]{000000} 0.74} & \cellcolor[HTML]{E58D94}{\color[HTML]{000000} 0.54} \\
obj=heat & \cellcolor[HTML]{F1F1F2}{\color[HTML]{000000} 0} & \cellcolor[HTML]{F1F1F2}{\color[HTML]{000000} 0} & \cellcolor[HTML]{F2F1F1}{\color[HTML]{000000} 0} & \cellcolor[HTML]{F2F1F1}{\color[HTML]{000000} 0} & \cellcolor[HTML]{F2F1F1}{\color[HTML]{000000} 0} & \cellcolor[HTML]{DD545E}{\color[HTML]{000000} 0.86} \\
loc=bedroom & \cellcolor[HTML]{F2EEEE}{\color[HTML]{000000} 0.02} & \cellcolor[HTML]{F2F1F1}{\color[HTML]{000000} 0} & \cellcolor[HTML]{F2EDED}{\color[HTML]{000000} 0.03} & \cellcolor[HTML]{F2EFEF}{\color[HTML]{000000} 0.01} & \cellcolor[HTML]{EDCCCE}{\color[HTML]{000000} 0.20} & \cellcolor[HTML]{DB404C}{\color[HTML]{000000} 0.97} \\ \hline
\end{tabular}
}
\caption{Example of paralinguistic explanation, \textsc{FSC} dataset, instance in Table \ref{tab-example-fsc-word}. The higher the value, the more the perturbations on the paralinguistic feature impact the prediction.}
\label{tab-example-fsc-context}
\end{table}

Table \ref{tab-example-fsc-word} shows the word-level audio segment explanation for this utterance computed with respect to the predicted class for each intent slot.
For each segment, we report its importance for the prediction.
We visualize only the word-level transcriptions for convenience and visualization constraints. However, recall that our approach works 
end-to-end at the audio level, and importance scores relate to audio segments.
The explanation reveals that the segment associated with the word `\textit{up}' is the most relevant term for the action \textit{increase}. 
Spoken words `\textit{heat}' and `\textit{bedroom}' are associated with the target object \textit{heat} and the target location \textit{bedroom}.
Hence, we can 
say that the explanation is \textit{plausible} and \textit{trust} the model for this prediction.

\begin{figure}[t]
\centering
\includegraphics[width=0.99\linewidth]{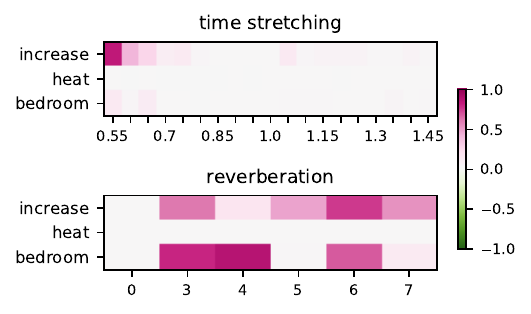}
\caption{
Heatmap of the prediction differences when varying the paralinguistic information. The higher the value, the more the paralinguistic changes impact the prediction.}
\label{fig-context-example-fsc-drilldown}
\end{figure}

Table \ref{tab-example-fsc-context} shows the paralinguistic explanation.
The prediction for this instance is greatly affected by the introduction of noise. The reverberation impacts the prediction for the slot action and slightly for the location; on the other hand, the object prediction is not affected.
The pitch transformation we introduce does not impact the predictions, both when increasing (`\textit{up}') and lowering (`\textit{down}') the pitch.
Finally, we reveal that shrinking the utterance duration (\textit{time} `\textit{stretch down}') and hence increasing the utterance speed impacts only the action \textit{increase}.

We can further inspect the impact of paralinguistic transformations on predictions by visualizing the prediction difference for each individual transformation via heatmaps.
Figure \ref{fig-context-example-fsc-drilldown} shows the prediction difference 
when stretching the audio and 
introducing reverberation.
Note that `1' and `0' are the reference values for time stretching and reverberation, respectively, and hence correspond to the 
original utterance.
We observe no impact when extending the utterance duration (values $\ge$1.05). 
At the same time, we note that the prediction probability of the action \textit{increase} highly changes when increasing the utterance speed (which corresponds to values 0.55-0.7). 

Our approach reveals the relevant factors for \textit{individual} predictions, and it is, hence, a tool for model understanding.
We include further examples of explanations in our repository.

\begin{figure}[!t]

\centering
\includegraphics[width=1\linewidth]{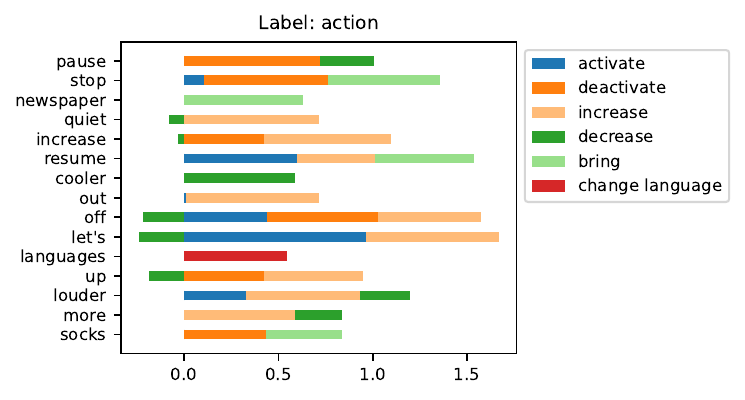}
\caption{Summary plot of average importance of word-level audio segments, separately for each predicted class. Top-15 segments, action label of \textsc{FSC} dataset.}
\label{fig-fsc-global-summary-action}
\end{figure}

\begin{table}[!t]
\centering
\setlength{\tabcolsep}{3pt}
\scalebox{0.92}{
\begin{tabular}{r|cc|cc|c|c}
\hline
\multicolumn{1}{l|}{}                   & \multicolumn{2}{c|}{\textbf{pitch}}                                                                       & \multicolumn{2}{c|}{\textbf{stretch}}                                                                     & \multicolumn{1}{c|}{}                                   & \multicolumn{1}{c}{}                                 \\
\multicolumn{1}{l|}{\multirow{-2}{*}{}} & \multicolumn{1}{c}{{\color[HTML]{000000} \textbf{down}}}        & \multicolumn{1}{c|}{{\color[HTML]{000000} \textbf{up}}}       & \multicolumn{1}{c}{{\color[HTML]{000000} \textbf{down}}}        & \multicolumn{1}{c|}{{\color[HTML]{000000} \textbf{up}}}       & \multicolumn{1}{c|}{\multirow{-2}{*}{\textbf{reverb}}} & \multicolumn{1}{c}{\multirow{-2}{*}{\textbf{noise}}} \\ \cline{2-7} 
{\color[HTML]{000000} action}           & \cellcolor[HTML]{F0E9F2}{\color[HTML]{000000} 0.04} & \cellcolor[HTML]{F1ECF2}{\color[HTML]{000000} 0.03} & \cellcolor[HTML]{ECD8EF}{\color[HTML]{000000} 0.13} & \cellcolor[HTML]{EEE0F0}{\color[HTML]{000000} 0.09} & \cellcolor[HTML]{E6BDEB}{\color[HTML]{000000} 0.27}     & \cellcolor[HTML]{D67BE2}{\color[HTML]{000000} 0.59}  \\
{\color[HTML]{000000} object}           & \cellcolor[HTML]{F1EEF2}{\color[HTML]{000000} 0.02} & \cellcolor[HTML]{F2EFF2}{\color[HTML]{000000} 0.01} & \cellcolor[HTML]{EFE3F1}{\color[HTML]{000000} 0.07} & \cellcolor[HTML]{F0E9F2}{\color[HTML]{000000} 0.05} & \cellcolor[HTML]{EACFEE}{\color[HTML]{000000} 0.17}     & \cellcolor[HTML]{D269DF}{\color[HTML]{000000} 0.69}  \\
{\color[HTML]{000000} location}         & \cellcolor[HTML]{F2EFF2}{\color[HTML]{000000} 0.01} & \cellcolor[HTML]{F2EFF2}{\color[HTML]{000000} 0.01} & \cellcolor[HTML]{F0E6F1}{\color[HTML]{000000} 0.06} & \cellcolor[HTML]{F1EBF2}{\color[HTML]{000000} 0.04} & \cellcolor[HTML]{EDDBF0}{\color[HTML]{000000} 0.11}     & \cellcolor[HTML]{E2ADE9}{\color[HTML]{000000} 0.35}  \\ \hline
\end{tabular}
}
\caption{Average paralinguistic attributions for the \textsc{FSC} dataset. The higher the score, the more the corresponding change in the paralinguistic feature impacts the prediction probability.}
\label{tab-fsc-summary-plot-context}
\end{table}

\begin{table*}[ht]
\centering
\setlength{\tabcolsep}{5pt}
\scalebox{0.95}{
\begin{tabular}{r|c|ccc|c|c}
\toprule
 &  & \multicolumn{3}{c|}{\textbf{FSC}} & \textbf{ITALIC} & \textbf{IEMOCAP} \\ \hline
\textbf{} & & \textbf{action} & \textbf{object} & \textbf{location} & \textbf{intent} & \textbf{emotion} \\ \midrule
\textsc{WA-L1O} & \multirow{2}{*}{\textit{Comprehensiveness}} & \textbf{0.619} & \textbf{0.623} & \textbf{0.465} & \textbf{0.693} & \textbf{0.508} \\
random &  & 0.294±0.005 & 0.246±0.003 & 0.195±0.006 & 0.324±0.005 & 0.273±0.005 \\ \hline
\textsc{WA-L1O} & \multirow{2}{*}{\textit{Sufficiency}} & \textbf{0.158} & \textbf{0.083} & \textbf{0.065} & \textbf{0.164} & \textbf{0.311} \\
random &  & 0.474±0.004 & 0.444±0.008 & 0.339±0.006 & 0.557±0.004 & 0.450±0.002 \\ \bottomrule
\end{tabular}
}
\caption{Comprehensiveness and Sufficiency results for our word attribution explanation via leave-one-out (\textsc{WA-L1O}) and random attribution for the \textsc{FSC}, \textsc{ITALIC}, and \textsc{IEMOCAP} datasets, separately for each label. For comprehensiveness, the closer to one, the better. For sufficiency, the closer to zero, the better.}
\label{tab-faithfulness}
\end{table*}

\paragraph{Global level.}
We can also analyze model behavior across the entire dataset.
We aggregate the importance scores of word audio segments or paralinguistic levels to investigate the \textit{global} influence of each component. 

Figure \ref{fig-fsc-global-summary-action} shows a summary plot for the word-level audio segment explanations of wav2vec 2.0 predictions on FSC test set for the label `action'.
We first compute the explanations for the predicted classes. Then, we aggregate audio segments corresponding to the same transcripted word after basic processing (i.e., lowercase and punctuation removal). We report the top 15 segments with the highest average importance. 
Each term represents the average importance scores separately for each class. 

The summary plot reveals which spoken words are associated with a predicted class.
From Figure \ref{fig-fsc-global-summary-action}, we infer that the importance scores for some spoken words such as `\textit{language}', `\textit{newspaper}', and `\textit{cooler}'  across the entire test set are associated with a single class value. 
Each class corresponds to a plausible value (`\textit{change language}', `\textit{bring}', and `\textit{decrease}'), making the explanations plausible.
In cases where a term is associated with multiple labels, the summary plot can serve as a debugging tool. 
For instance, the spoken word `\textit{pause}' is correctly linked to the predicted action `\textit{deactivate}' but erroneously connected to `\textit{decrease}'. 
Similar considerations apply to the other two labels we include in the repository.

Table \ref{tab-fsc-summary-plot-context} shows the average importance score of paralinguistic explanations aggregated for each label. The results reveal that adding background noise globally impacts the model prediction. The reverberation affects more the predictions of the action label than the ones of the location. 
We observe higher average importance scores for the action label for the time stretching component, specifically when compressing the utterance duration (`\textit{stretch down}') and, therefore, increasing the audio speed. Conversely, the pitch transformation we introduce generally does not impact the predictions.

\subsection{Quantitative evaluation}

In this section, we quantitatively evaluate the quality of our explanations. 
A critical requirement for explanations is their faithfulness to the model. 
Faithfulness measures evaluate how accurately the explanation reflects the model's inner workings \citep{jacovi-goldberg-2020-towards}.


\paragraph{Metrics.} We generalize two widely adopted measures for the XAI literature: \textit{comprehensiveness} and \textit{sufficiency} \citep{deyoung-etal-2020-eraser}. These notions were originally designed for token-level explanations for text classification, where explainers assign a relevance score to each token. This scenario is close to our word-level audio segment explanations. Intuitively, we consider audio segments rather than tokens. 
\textit{Comprehensiveness} evaluates whether the explanation captures the audio segments the model used to make the prediction. We measure it by progressively masking the audio segments highlighted by the explanation, observing the change in probability, and finally averaging the results. A high value of comprehensiveness indicates that the audio segments highlighted by the explanations are relevant to the prediction.
Conversely, \textit{sufficiency} captures if the audio segments in the explanation are sufficient for the model to make the prediction. Opposed to comprehensiveness, we preserve only the relevant audio segments and compute the prediction difference. A low score indicates that the audio segments in the explanations indeed drive the prediction. We include the extended description of the two metrics in our repository.

\paragraph{Baseline.} 
We assess the quality of explanations compared to a random explainer.
The random explainer assigns a random score in the range [-1, 1] to each word audio level segment.

\paragraph{Results.}
Table \ref{tab-faithfulness} shows the comprehensiveness and sufficiency results on the \textsc{FSC}, \textsc{ITALIC}, and \textsc{IEMOCAP} datasets, separately for each label. 
We generate our word-level explanations with respect to the predicted class.
For the random baseline, we consider five rounds of generations, and we report average and standard deviation.
The results show that our word-level audio segment explanations computed by leaving one out audio segments (\textsc{WA-L1O} in Table \ref{tab-faithfulness}) highly outperform the random baseline for both metrics.

\section{Related Work}
\label{sec:rw}

\subsection{Interpretability for Speech Models}

Multiple studies have adopted Layer-wise Relevance Propagation (LRP) \citep{bach2015pixel}, initially proposed for image classification explanations, to explain prediction across diverse audio analysis tasks. 
Most of these works represent explanations as time-frequency heatmaps over spectrograms, such as \citet{becker2018interpreting} for gender and digit audio classification, \citet{frommholz2023xai} for audio event classification, and \citet{colussi2021interpreting} for the task of urban sound classification.
\citet{wang2023explainable} used heatmaps over ad-hoc terms (carrier and modulation frequency) for the specific task of audio classification of playing techniques (e.g., vibrato, trill, tremolo) in the context of music signal analysis.
While experts can find spectrograms a familiar tool for understanding audio data, these visual representations can be challenging for laypersons to interpret. 

\citet{becker2018interpreting} also adopt the LRP method to derive the relevance score of individual samples with respect to the input waveform in the time domain.
Interpreting explanations as sets of individual samples can pose challenges, such as a lack of abstraction and context of isolated data points.
We advocate for prioritizing a more user-friendly and intuitive approach to explanation. 
In this line of intent, rather than samples, \citet{wu2023explanations} assign relevance scores to audio frames, i.e., raw data bins in time dimension of predefined size. The work generalizes two XAI techniques from image classification and explains Automatic Speech Recognition (ASR) systems. 
\citet{mishra2017local} propose to describe the data to explain via interpretable representations. Their method involves segmenting the data into equal-width segments within the time, frequency, or time-frequency domains. Subsequently, they apply the LIME explanation method \citep{ribeiro2016should} to these interpretable representations.
However, these temporal explanations may be affected by the size of the audio segments chosen for analysis. 
Moreover, they are not grounded in spoken words or paralinguistic information, hindering interpretability for semantically intensive contexts such as speech classification.

The work by \citet{wu2023can} aligns with our direction, as it not only tests fixed-width audio segments but also audio segments aligned with phonemes. 
However, the approach requires phoneme-level annotations, and therefore, it is limited to evaluation purposes when such labeling is available. Moreover, the method is suitable for the phoneme recognition task. 
In contrast, our approach offers a more generalized solution to any Speech Language Understanding (SLU) classification model and data. We automatically derive audio segments at the word level, coupled with their transcriptions, via state-of-the-art speech transcription systems. 
Furthermore, our approach stands out as the first to offer explanations that study the impact of paralinguistic features on predictions, presenting these insights in an interpretable form.

\subsection{Explanation by Occlusion}

Removing parts of input data to understand their impact is a well-established strategy in explainability \citep{covert2021explaining}. Different domains use various techniques for removing or masking parts of the data. Standard techniques for image data include noise addition, blurring, or replacing via a grey area. Using a special mask token or directly removing words is often employed in text analysis. For structured data, analyzing the effects based on average values is a typical approach~\citep{covert2021explaining}.
For speech data, \citet{wu2023can} have applied a similar technique to phonemes, using signal zeroing for masking. However, the masking is adopted for generating perturbation used by LIME explanation method \citep{ribeiro2016should}, and they are at the phoneme level.

\section{Conclusion}
\label{sec:conclusion}

We propose a novel perturbation-based explanation method that explains the predictions of speech classification models 
regarding word-level audio segments and paralinguistic features.
Our results show that our explanations can be a tool for model understanding.

\section*{Limitations}

Our work has some technical and design limitations. From the technical perspective, word-level segment attributions are computed by masking one-word segment at a time, thus not considering the intersectional effect of multiple masked words. We plan to experiment with different masking strategies.
Moreover, word-level explanations might not be the most helpful explanation in specific speech classification tasks, e.g., spoken language identification or speaker identification. We are accounting for this limitation by including paralinguistic explanations, but we will also explore new methods.
We will also investigate the impact of the perturbation techniques and third-party speech libraries on paralinguistic explanations.
From the experimental design perspective, we are currently reporting self-evaluation for plausibility. We will conduct a comprehensive user study to evaluate it thoroughly.




\bibliographystyle{acl_natbib}

\bibliography{anthology,custom}

\end{document}